\documentclass[letterpaper]{article} 
\usepackage{aaai2026}  
\nocopyright
\usepackage{times}  
\usepackage{helvet}  
\usepackage{courier}  
\usepackage[hyphens]{url}  
\usepackage{graphicx} 
\usepackage{natbib}  
\usepackage{caption} 
\usepackage{amsmath}
\usepackage{amssymb}
\usepackage{booktabs}

\title{Training-Free Halving of Activated Experts\\in Fine-Grained Mixture-of-Experts Models}

\author{
    Xing Chen\footnote{Correspondence: raincchio@gmail.com},
    Hengshuai Yao
}

\begin{document}

\maketitle

\begin{abstract}
Modern fine-grained Mixture-of-Experts (MoE) models route each token to a small
number of experts and renormalize their router probabilities. We show that this
renormalization implicitly calibrates expert output gain to the training top-$k$:
reducing $k$ at inference changes not only which experts are used but also the
strength of the expert branch. We separate these effects by activating the top $k_1$
experts while normalizing by the probability mass of the top $k_2$ experts,
introducing one integer with no parameters, training, or measurable compute overhead.
On Qwen3.6-35B-A3B, reducing from 8 to 4 experts causes a 4.65-point MMLU drop under
standard renormalization but only 0.35 points with $k_2=16$, while halving
routed-expert compute. The result replicates on the $11\times$ larger
Qwen3.5-397B-A17B, where reducing from 10 to 5 experts loses only 0.55 points with an
appropriate reference set. Removing renormalization entirely is catastrophic, showing
that preserving a suitable reference mass is crucial. We further find that perplexity
and downstream accuracy favor different $k_2$, cautioning against selecting MoE
compression settings using unlabeled text alone. Analyses also show that expert
identity matters substantially more than expert weighting, while balanced and
domain-specialized routing leaves limited room for expert pruning.
\end{abstract}

\section{Introduction}

Sparse Mixture-of-Experts (MoE) architectures decouple parameter count from
per-token compute by activating only a few of many expert
sub-networks~\citep{shazeer2017outrageously,lepikhin2021gshard,fedus2022switch}.
Recent open-weight models have pushed this design toward \emph{fine-grained}
sparsity: rather than 8 or 16 large experts, they use hundreds of small ones and
route each token to a handful of them~\citep{dai2024deepseekmoe,deepseekv3,qwen3}.
In Qwen3.6-35B-A3B, our primary object of study, all 40 layers are MoE layers with
256 experts each and $k{=}8$ selected per token; the routed experts hold $32.2$B of
the model's $35.9$B parameters (89.6\%), while each token touches only $3.1\%$ of them.

This memory--compute asymmetry---all experts must be resident, few are used---is what
makes MoE compression attractive and awkward at the same time. Reducing the number of
experts (pruning or merging) saves memory but not compute, since $k$ is unchanged;
reducing $k$ saves compute but not memory. A large body of training-free work
attacks the memory side by identifying and removing redundant
experts~\citep{lu2024notallexperts,li2024mcsmoe}. The compute side, by contrast, is
usually treated as trivially adjustable: $k$ is an inference-time argument, so one
simply lowers it, renormalizes over the survivors, and measures the
loss~\citep{chittyvenkata2025lexi}. The loss is large, and the reported remedies are
training-based---retraining with sampled expert counts, or distilling a
half-expert student~\citep{wang2025matryoshkamoe,gu2025elasticmoe,lv2026skiphalf}.

We argue this treatment overlooks a detail that turns out to dominate the outcome.
Nearly all modern MoE implementations renormalize the selected router probabilities,
$w_i = p_i / \sum_{j \in \mathcal{T}_k} p_j$, so that the mixture weights sum to one.
In a coarse-grained router this is nearly a no-op, because the top few probabilities
already carry most of the mass. In a fine-grained router it is not: in the model we
study, the top-8 of 256 probabilities sum to only $0.182$, so renormalization
multiplies them by $5.5$ on average. That amplification is not a free-standing design
choice---it is a quantity the model was trained under, and it is a function of $k$.
Lowering $k$ from 8 to 4 while keeping the renormalization discards four experts
\emph{and} redistributes their weight onto the survivors, inflating the expert
branch's contribution to the residual stream beyond anything the model saw in
training. The measured degradation of ``top-4'' therefore conflates two effects: the
loss of four experts, and a gain miscalibration that has nothing to do with expert
capacity.

Our contribution is to separate them, using the observation that the numerator and
the denominator of the renormalization need not use the same expert set. We activate
the top $k_1$ experts but normalize by the mass of the top $k_2$:
\[
w_i = \frac{p_i}{\sum_{j \in \mathcal{T}_{k_2}} p_j},
\qquad i \in \mathcal{T}_{k_1},
\]
where $k_1$ controls compute and $k_2$ controls gain. Setting $k_2{=}k_1$ recovers
standard renormalization; $k_2{=}E$ removes it; $k_2{=}k$ anchors the gain to the
trained value; and $k_2 > k$ pushes the gain below its trained value, adaptively per
token. Only $k_1$ expert FFNs are ever evaluated, and the router already computes all
$E$ probabilities, so varying $k_2$ costs nothing measurable.

The empirical payoff is large, and it is largest exactly where prior practice reports
the reduction of $k$ as unaffordable. On 2000-question, 5-shot MMLU with paired
McNemar testing, activating 4 of 256 experts per layer instead of 8---halving
routed-expert compute---costs $4.65$ accuracy points under standard renormalization
($p{=}1.7{\times}10^{-9}$), but only $0.35$ points at $k_2{=}16$ ($p{=}0.66$), which
the paired test cannot distinguish from the top-8 baseline. The same experiment on
Qwen3.5-397B-A17B, an $11\times$ larger model with 512 experts and $k{=}10$, gives
$2.10$ points at $k_2{=}k_1{=}5$ versus $0.55$ points at $k_2{=}10$ ($p{=}0.24$).

We summarize our findings as follows.

\begin{itemize}
\item \textbf{The renormalization reference set is a free variable.} Lowering $k$
without decoupling the denominator conflates capacity loss with gain miscalibration.
Decoupling it is a one-line change that makes it possible to halve the number of
activated experts with no statistically detectable MMLU loss on two models spanning
$11\times$ in scale.
\item \textbf{The gain, not the weights, is what breaks.} Removing renormalization
entirely ($k_2{=}E$) costs $27.5$ MMLU points and $356\%$ perplexity, so the model is
not indifferent to weight magnitudes; and with $k_1$ held fixed, both shrinking and
enlarging the reference set away from its optimum degrade the model. The damage is
specific to the reference set being wrong, and it is two-sided.
\item \textbf{Which experts, not how much.} With the number of activated experts held
fixed at 8, replacing the router's weights with uniform ones costs $+49.7\%$
perplexity, while sampling experts instead of taking the argmax costs $+310\%$ and
random selection $+1667\%$. Expert \emph{identity} is worth several times more than
expert \emph{weighting}, which rules out approximate or stochastic routing on this
model but is precisely what makes a cheap global gain correction viable.
\item \textbf{Perplexity picks the wrong operating point.} On the 35B model
perplexity is minimized at $k_2{=}8$ and MMLU at $k_2{=}16$; the perplexity-optimal
setting is significantly worse on MMLU ($-1.10$ points, $p{=}0.021$). Selecting
compression hyperparameters on unlabeled text is not safe here, and we recommend
paired significance testing on a downstream task.
\item \textbf{Expert pruning has less headroom than this.} The router is well balanced
with essentially no dead experts, and its specialization is strongly domain-dependent:
WikiText and code top-64 expert sets overlap at $0.164$, below the $0.250$ random
baseline, so domain-pruned models do not transfer and the union of two domains'
top-128 sets already covers 198 of 256 experts.
\end{itemize}

All results are obtained without any training, distillation, or fine-tuning.

\section{Related Work}

\paragraph{Sparse MoE language models.}
Conditional computation via learned routing dates to~\citet{shazeer2017outrageously},
and was scaled to modern transformers by GShard~\citep{lepikhin2021gshard} and Switch
Transformer~\citep{fedus2022switch}, which also documented the training instabilities
that motivate auxiliary load-balancing losses~\citep{zoph2022stmoe}. Alternatives to
token-choice top-$k$ routing include expert-choice routing~\citep{zhou2022expertchoice}.
Open-weight decoder MoEs such as Mixtral~\citep{jiang2024mixtral} use coarse
granularity (8 experts, top-2), whereas DeepSeekMoE~\citep{dai2024deepseekmoe} argued
for many small experts plus always-on shared experts, a design adopted by
DeepSeek-V3~\citep{deepseekv3} and the Qwen3 series~\citep{qwen3}, and pushed further
by~\citet{he2024mixtureofamillionexperts}. Our objects of study belong to this
fine-grained family, and the effect we describe is a direct consequence of fine
granularity: it is the flatness of a 256-way router's top-$k$ mass that makes the
renormalization gain large and thus makes miscalibrating it expensive.

\paragraph{Analyses of learned routing.}
Several studies have asked whether experts specialize semantically. OpenMoE~\citep{xue2024openmoe}
reported context-independent, largely token-identity-driven routing, and~\citet{lo2024closer}
analyzed routing behavior across layers in open MoE models. Our routing measurements
are consistent with the general picture that routers are well balanced, but we reach
a sharper conclusion on specialization for this model: the domain effect exceeds a
within-domain resampling noise floor by nearly four orders of magnitude, and
cross-domain expert overlap falls \emph{below} chance. We also separate layers by
attention type---both models interleave linear-attention and full-attention layers in
a 3:1 pattern---and find the two classes behave differently, a grouping that prior
analyses of homogeneous architectures had no reason to consider.

\paragraph{Training-free MoE compression.}
Expert pruning removes experts judged redundant~\citep{lu2024notallexperts} and
merging consolidates them using routing statistics~\citep{li2024mcsmoe};
\citet{lasby2025reap} report near-lossless one-shot pruning at 50\% expert reduction
with a router-weighted criterion. These target memory, as does
quantization~\citep{frantar2023gptq,lin2024awq}, which suits MoE well since the routed
experts are a large homogeneous block; \citet{huang2025mixturecompressor} combine both
in a training-free pipeline. On the compute side the closest method to ours is
LExI~\citep{chittyvenkata2025lexi}, which picks a per-layer number of active experts
data-free---and renormalizes the surviving router weights, which in our
parameterization is $k_2{=}k_1$, precisely the setting we find worst on both models.
The two compose: LExI decides \emph{how many} experts a layer keeps, the reference set
decides \emph{how strongly} the survivors drive the residual stream. More generally,
we are not aware of prior work that scans the renormalization denominator separately
from $k$, and prior top-$k$ reduction numbers on fine-grained models are, for this
reason, likely to be pessimistic. Some architectures do expose a related
knob---GLM-style configurations ship a \texttt{routed\_scaling\_factor} applied after
renormalization---but as a fixed scalar chosen at training time.

\paragraph{Changing $k$ after training.}
That MoE models degrade sharply when the number of activated experts is altered at
inference is well documented---\citet{wang2025matryoshkamoe} call the degradation
``precipitous''---and the reported remedies are \emph{training-based}: retraining with
randomly sampled expert counts~\citep{wang2025matryoshkamoe}, training expert
combinations so that $k$ can be scaled \emph{up}~\citep{gu2025elasticmoe}, null experts
that let the effective $k$ vary per token~\citep{zeng2024adamoe}, and, closest to our
headline result, self-distilling a post-trained MoE that skips roughly half its
experts~\citep{lv2026skiphalf}. That last work also argues, as we do, that
renormalizing the surviving weights inflates the effective scale of the expert residual
branch relative to what pre-training calibrated. Our contribution relative to it is
twofold: our correction needs no training, and we find the choice is not the binary
``renormalize or do not''---those are the endpoints $k_2{=}k_1$ and $k_2{=}E$ of a
family whose useful settings are interior. Which endpoint is less wrong depends on the
model's native forward pass: for a model that does not renormalize, not renormalizing
preserves the trained gain, whereas both models here renormalize by construction and
removing it costs $27.45$ MMLU points. The invariant is not a rule about
renormalization but the reference set.

\section{Background and Setup}

\paragraph{Models.}
We study two fine-grained MoE models from the same series but very different scales
(Table~\ref{tab:models}). Qwen3.6-35B-A3B has 40 layers, hidden size 2048, and 256
routed experts of intermediate size 512 per layer with $k{=}8$, plus one always-on
shared expert per layer modulated by a learned scalar gate; every layer is an MoE
layer. Qwen3.5-397B-A17B has 60 layers, hidden size 4096, and 512 routed experts of
intermediate size 1024 with $k{=}10$. Both interleave linear-attention (gated
delta-net style) and full-attention blocks in a repeating 3:1 pattern. For the 35B
model we enumerated tensors in the released checkpoint: routed experts hold 89.6\% of
all parameters ($32.212$B of $35.952$B; the remainder is $2.448$B of attention,
embeddings, shared experts and norms, a $0.845$B multi-token-prediction head, and a
$0.447$B vision tower). Each token activates $8{+}1$ of 256 experts per layer, i.e.
$3.1\%$ of expert parameters---a $32{:}1$ memory-to-compute asymmetry that organizes
the rest of this paper.

\begin{table}[t]
\centering
\footnotesize
\setlength{\tabcolsep}{4pt}
\begin{tabular}{lrr}
\toprule
& Qwen3.6 & Qwen3.5 \\
& 35B-A3B & 397B-A17B \\
\midrule
Layers (all MoE)      & 40    & 60 \\
Hidden size           & 2048  & 4096 \\
Experts per layer $E$ & 256   & 512 \\
Trained $k$           & 8     & 10 \\
Expert interm.\ size  & 512   & 1024 \\
Shared experts        & 1     & 1 \\
Top-$k$ router mass   & 0.182 & 0.192 \\
Renorm.\ gain         & $5.50\times$ & $5.22\times$ \\
\midrule
Act.\ expert params, native  & 1.132\,B & 8.305\,B \\
\quad at $k_1 = 0.75k$/$0.8k$ & 0.881\,B & 6.795\,B \\
\quad at $k_1 = 0.5k$         & 0.629\,B & 4.530\,B \\
\bottomrule
\end{tabular}
\caption{The two models. ``Act.\ expert params'' counts routed experts activated per
token plus the always-on shared expert, summed over layers. Router mass and gain are
measured on WikiText.}
\label{tab:models}
\end{table}

\paragraph{Router and the implicit gain.}
For hidden state $x$ at a given layer, the router computes $p = \mathrm{softmax}(W_g x)$
over all $E$ experts, selects the index set $\mathcal{T}_k$ of the $k$ largest
entries, and forms the layer output
\begin{equation}
y = \sum_{i \in \mathcal{T}_k} w_i\, E_i(x) + g_{\mathrm{sh}}\, E_{\mathrm{sh}}(x),
\qquad
w_i = \frac{p_i}{\sum_{j \in \mathcal{T}_k} p_j}.
\label{eq:native}
\end{equation}
The denominator is what concerns us. Write $m_k(x) = \sum_{j \in \mathcal{T}_k} p_j$
for the retained probability mass. Renormalization multiplies the raw probabilities by
$1/m_k(x)$, and since $m_k$ increases with $k$, this gain is implicitly tied to the
$k$ used in training. We measure $\mathbb{E}[m_8] = 0.182$ on WikiText and $0.170$ on
code for the 35B model, i.e. average amplifications of $5.50\times$ and $5.87\times$;
for the 397B model $\mathbb{E}[m_{10}] = 0.192$, i.e. $5.22\times$. Dropping to
$k{=}4$ reduces $m_k$ and hence \emph{raises} the gain applied to the surviving
experts, pushing the expert branch out of the regime the model was trained in. This is
the miscalibration we correct.

\paragraph{Method: decouple the reference set.}
We keep the selection rule and the numerator, and replace only the denominator's index
set:
\begin{equation}
w_i = \frac{p_i}{\sum_{j \in \mathcal{T}_{k_2}} p_j},
\qquad i \in \mathcal{T}_{k_1},
\label{eq:ours}
\end{equation}
with $\mathcal{T}_{k_1} \subseteq \mathcal{T}_{k_2}$ when $k_2 \ge k_1$. The
parameterization is a strict generalization of common practice:
\begin{itemize}\itemsep2pt
\item $k_2 = k_1$: standard renormalization (Eq.~\ref{eq:native});
\item $k_2 = k$ (trained value): the gain is anchored to what training calibrated;
\item $k_2 = E$: no renormalization at all, since $\sum_j p_j = 1$;
\item $k_2 > k$: gain below the trained value, applied per token.
\end{itemize}
The weights no longer sum to one when $k_2 > k_1$; they sum to
$m_{k_1}(x)/m_{k_2}(x) < 1$, which is a \emph{per-token, input-adaptive}
down-scaling rather than a constant. Averaged over tokens this ratio is $0.854$ for
$(k_1,k_2){=}(6,8)$ and $0.592$ for $(6,16)$ on the 35B model, so the discrete grid we
scan spans the range a tuned global scalar would explore, without introducing a
continuous hyperparameter.

Only $k_1$ expert FFNs are evaluated, so the compute saving is exactly that of
reducing $k$ to $k_1$. The extra cost of a larger $k_2$ is a wider top-$k$ over router
logits that are computed for all $E$ experts anyway---negligible against $k_1$ expert
FFNs. Both are implemented by replacing the router's \texttt{forward} at inference
time; no weights are modified.

\paragraph{Evaluation protocol.}
Perplexity is measured on 100K-token samples of WikiText-103~\citep{merity2017pointer}
and a Python subset of CodeParrot~\citep{tunstall2022codeparrot}, at a 2048-token
context, with a held-out WikiText
split (drawn beyond row 400{,}000, disjoint from the tuning split) used to check that
the choice of $k_2$ does not overfit. All configurations are scored on byte-identical
token chunks. Downstream accuracy is measured on MMLU~\citep{hendrycks2021mmlu}: 2000
questions sampled with a fixed seed from all 14{,}042 questions across all 57
subjects, 5-shot, scored by comparing the logits of the single tokens
\texttt{A}--\texttt{D}; on C-Eval~\citep{huang2023ceval} (1300 questions, 5-shot,
Chinese); and, as a generation task, on GSM8K~\citep{cobbe2021gsm8k} (500 questions,
0-shot, greedy to 512 tokens, reasoning disabled via the chat template, exact-match on
the final number). All configurations run on \emph{identical} question sets, compared to
the baseline with a paired McNemar test~\citep{mcnemar1947note}. Pairing is not optional
at these effect sizes: a power analysis at $n{=}2000$ gives a minimum detectable
difference of $0.98\%$ paired versus $3.04\%$ unpaired, and our effects are $0.3$--$5\%$.

\section{How the Router Actually Behaves}

Before modifying the router we characterize it, using forward hooks on all gates to
record the selected expert indices and softmax weights over 500K tokens per domain
(WikiText and code), giving an expected 15{,}680 activations per expert per layer on
the 35B model.

\begin{figure*}[t]
\centering
\includegraphics[width=0.86\textwidth]{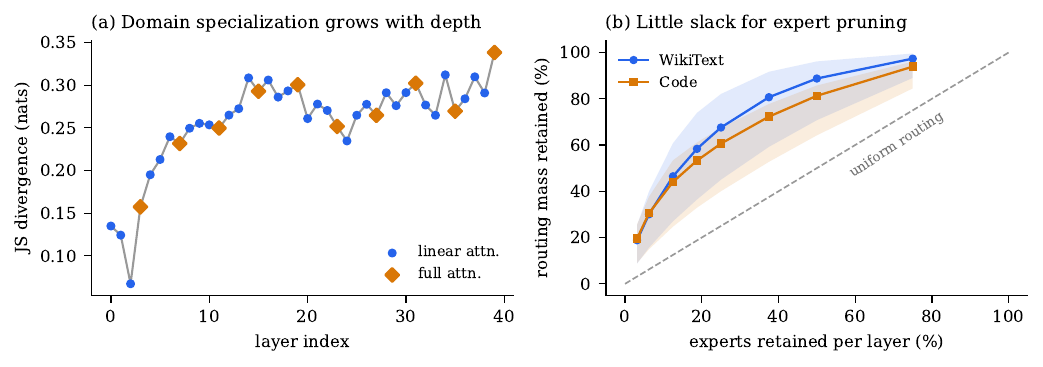}
\caption{Routing structure of Qwen3.6-35B-A3B measured over 500K tokens per domain.
(a) Cross-domain (WikiText vs.\ code) Jensen--Shannon divergence of the per-layer
expert distribution rises with depth; full-attention layers are marked separately.
(b) Fraction of routing mass retained when keeping only the top-$N$ experts per layer;
shaded bands span the best and worst layer. Even keeping half the experts loses
11--19\% of routing mass, and the worst layer far more.}
\label{fig:routing}
\end{figure*}

\paragraph{Load is well balanced; there is no obvious fat to cut.}
Normalized activation entropy is $0.889$ (WikiText) and $0.915$ (code), with Gini
coefficients of $0.581$ and $0.492$. The number of never-activated experts per layer
has median $0$ and maximum $1$ across all layers and both domains. The 397B model
behaves similarly (entropy $0.911$, Gini $0.527$, dead experts median $1$ of 512).
Figure~\ref{fig:routing}(b) shows the resulting retention curves: keeping the top
50\% of experts per layer retains only $88.8\%$ (WikiText) and $81.3\%$ (code) of
routing mass, and the worst layer retains just $64.2\%$. A router trained with
load-balancing pressure leaves little slack for pruning, which is the first reason we
look elsewhere for compute savings.

\paragraph{Experts are strongly domain-specialized.}
Comparing per-layer expert distributions across domains yields a mean Jensen--Shannon
divergence of $0.258$ nats. To calibrate this we estimate a noise floor by
multinomial resampling within a domain at matched counts, obtaining
$3.2\times10^{-5}$ nats; the observed divergence is roughly $8{,}200\times$ the floor,
so it is not a sampling artifact. More strikingly, the top-64 expert sets of the two
domains overlap at $0.164$, \emph{below} the random baseline of $0.250$: the domains
do not merely prefer different experts, they actively avoid each other's.
Specialization increases with depth (Figure~\ref{fig:routing}(a)), from $0.135$ nats
at layer 0 to $0.338$ at layer 39.

\paragraph{But specialization does not make pruning transferable.}
Pruning to the top-128 experts using code-derived rankings retains $81.3\%$ of code
routing mass; using WikiText-derived rankings retains only $38.0\%$ on the same
domain---a factor of two. Meanwhile the union of the two domains' top-128 sets covers
198 of 256 experts. Strong specialization is thus a double-edged result: it licenses
\emph{domain-specific} pruned models, but it forecloses a single general-purpose
pruned model, because any expert set broad enough to serve both domains is barely
smaller than the original.

\paragraph{Attention type structures routing.}
Grouping layers by attention type reveals structure that a layer-agnostic average
hides. Full-attention layers share expert preferences with each other (cross-layer
top-64 overlap $0.309$ versus the $0.250$ random baseline), whereas linear-attention
layers are mutually near-independent ($0.258$), and cross-type pairs are
indistinguishable from chance ($0.251$). The shared-expert gate follows the same
split: mean $0.178$ on full-attention layers versus $0.151$ on linear-attention
layers, rising with depth to $0.405$ at layer 39. We report this as a caution for
routing analyses of the increasingly common hybrid-attention architectures: layers are
not interchangeable samples.

\section{Expert Selection Matters More Than Expert Weighting}

If a global correction to the router weights is to work, the model must be more
sensitive to expert selection than to the precise weight values. We test this directly
with three ablations that all keep the number of activated experts at exactly 8, so
compute is held constant (Table~\ref{tab:ablation}).

\begin{table}[t]
\centering
\small
\begin{tabular}{lrrrr}
\toprule
& \multicolumn{2}{c}{WikiText} & \multicolumn{2}{c}{Code} \\
\cmidrule(lr){2-3}\cmidrule(lr){4-5}
Routing rule & PPL & $\Delta$ & PPL & $\Delta$ \\
\midrule
top-$k$ (native)      & 6.64   & ---      & 2.59  & --- \\
top-$k$, uniform $w$  & 9.95   & +49.7\%  & 3.11  & +20.3\% \\
sampled without repl. & 27.22  & +310\%   & 6.09  & +135\% \\
uniformly random      & 117.39 & +1667\%  & 18.69 & +623\% \\
\bottomrule
\end{tabular}
\caption{Routing ablations at fixed compute (8 experts activated in every row) on
Qwen3.6-35B-A3B. Replacing the router's weights with uniform ones is far less
damaging than replacing its \emph{choices}.}
\label{tab:ablation}
\end{table}

Discarding the router's weights entirely but keeping its choices costs $+49.7\%$
perplexity on WikiText. Keeping the router's distribution but sampling from it instead
of taking the argmax costs $+310\%$, and choosing experts uniformly at random costs
$+1667\%$. Selection is worth roughly six times more than weighting on both
domains.

This has two consequences. Negatively, it rules out a family of efficiency methods on
this model: approximate top-$k$, locality-sensitive-hashing routers, and any scheme
that injects stochasticity into selection will be catastrophic, because exact argmax
is effectively a hard requirement. Positively, it is what makes our method plausible.
Since the model tolerates substantial distortion of the weight \emph{magnitudes} as
long as the selected set is right, correcting the systematic component of that
distortion is enough---and reducing $k_1$ produces exactly a systematic, not
idiosyncratic, magnitude error.

\section{The Reference Set Controls the Damage}

\paragraph{The router's top-$k$ mass is small.}
Table~\ref{tab:mass} reports the mean softmax mass by rank on the 35B model. The single
highest-scoring expert receives only $0.046$ of the probability mass, the top-8
together only $0.182$, and even the top-32 only $0.380$. The distribution is remarkably
flat, as one would expect from a 256-way router trained with load balancing.
Renormalization therefore performs a $5.50\times$ amplification, and that factor is a
property of $k{=}8$.

\begin{table}[t]
\centering
\small
\setlength{\tabcolsep}{3.8pt}
\begin{tabular}{lrrrrr}
\toprule
Domain & rank 1 & top-4 & top-6 & top-8 & top-16 \\
\midrule
WikiText & 0.046 & 0.122 & 0.155 & 0.182 & 0.262 \\
Code     & 0.046 & 0.116 & 0.146 & 0.170 & 0.245 \\
\bottomrule
\end{tabular}
\caption{Cumulative router softmax mass by rank on Qwen3.6-35B-A3B, averaged over
tokens and layers. The top-8 of 256 experts carry under a fifth of the total mass, so
the renormalization in Eq.~\ref{eq:native} applies a large, $k$-dependent gain.}
\label{tab:mass}
\end{table}

\begin{figure*}[t]
\centering
\includegraphics[width=0.86\textwidth]{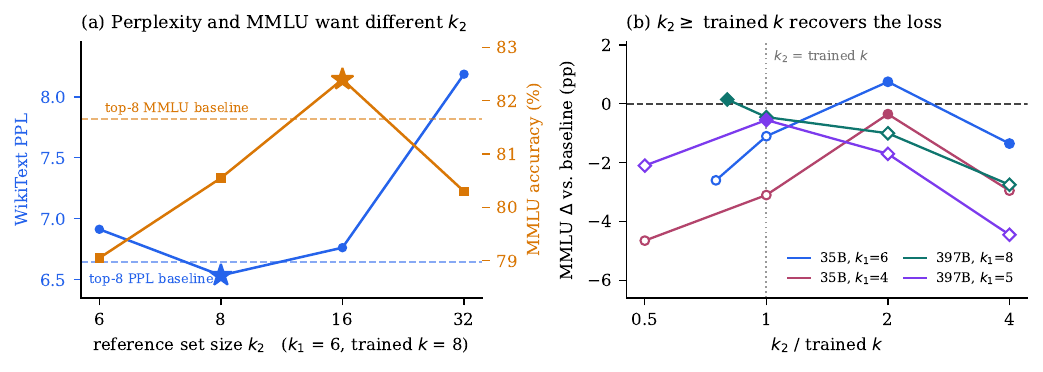}
\caption{(a) WikiText perplexity (blue, left) and MMLU accuracy (orange, right) as a
function of the reference set size $k_2$ at $k_1{=}6$ on Qwen3.6-35B-A3B; stars mark
the optima, dashed lines the native top-8 baselines. The two metrics disagree:
perplexity is minimized at $k_2{=}8$, MMLU at $k_2{=}16$. (b) MMLU change versus
$k_2/k$ for both models at two values of $k_1$ each; hollow markers are significant
at $p<0.05$ under a paired McNemar test, filled markers are not. Shrinking the
reference set together with $k_1$ (leftmost point of each curve) is significantly
harmful in three of four settings; $k_2 \ge k$ recovers it. The $k_2{=}E$ endpoint is
off-scale and omitted.}
\label{fig:s}
\end{figure*}

\paragraph{Both extremes of the reference set are harmful.}
Two observations bracket the effect. Removing renormalization altogether
($k_2{=}E{=}256$) at $k_1{=}6$ raises WikiText perplexity from $6.64$ to $30.29$
($+356\%$) and drops MMLU by $27.45$ points; the model depends on the amplification
rather than merely tolerating it. At the other extreme, shrinking the reference set
along with $k_1$ ($k_2{=}k_1$) costs $+4.08\%$ perplexity and $2.60$ MMLU points at
$k_1{=}6$, and $4.65$ MMLU points at $k_1{=}4$. Between them, $k_2$ in the range
$[k, 2k]$ recovers nearly all of it. Perplexity as a function of $k_2$ is thus
U-shaped with an interior minimum, and MMLU is inverted-U with an interior maximum
(Figure~\ref{fig:s}); neither optimum sits at an endpoint of the scanned grid.

\paragraph{Perplexity.}
Table~\ref{tab:ppl} gives the perplexity scan. On the 35B model at $k_1{=}6$, standard
renormalization costs $+4.08\%$, while $k_2{=}8$ yields $-1.61\%$---slightly
\emph{better} than the native top-8 baseline while activating 22\% fewer expert
parameters. The held-out WikiText split, disjoint from the split used to choose $k_2$,
reproduces the ranking exactly ($-0.57\%$ at $k_2{=}8$), as does code ($+0.68\%$),
so the choice is not an artifact of the tuning split. The 397B model shows the same
ordering at $k_1{=}8$, with $k_2{=}10$ the best of the grid on all three corpora
(WikiText, held-out, and code). Across both models and all six corpus/model
combinations, perplexity is minimized at $k_2$ equal to the model's trained $k$ and
degrades monotonically as $k_2$ moves away in either direction.

\begin{table}[t]
\centering
\small
\begin{tabular}{lrrr}
\toprule
Configuration & WikiText & Held-out & Code \\
\midrule
\multicolumn{4}{l}{\emph{Qwen3.6-35B-A3B}\quad(trained $k{=}8$, $E{=}256$)} \\
$k_1{=}8$ native   & 6.643 & 6.986 & 2.585 \\
$k_1{=}6$, $k_2{=}6$   & +4.08\% & +4.69\% & +1.72\% \\
$k_1{=}6$, $k_2{=}8$   & \textbf{--1.61\%} & \textbf{--0.57\%} & \textbf{+0.68\%} \\
$k_1{=}6$, $k_2{=}16$  & +1.81\% & +1.42\% & +3.29\% \\
$k_1{=}6$, $k_2{=}32$  & +23.25\% & +19.04\% & +13.00\% \\
$k_1{=}6$, $k_2{=}256$ & +356\% & +313\% & +158\% \\
\midrule
\multicolumn{4}{l}{\emph{Qwen3.5-397B-A17B}\quad(trained $k{=}10$, $E{=}512$)} \\
$k_1{=}10$ native  & 4.159 & 3.198 & 2.249 \\
$k_1{=}8$, $k_2{=}8$   & +1.06\% & +1.54\% & +0.42\% \\
$k_1{=}8$, $k_2{=}10$  & \textbf{+0.46\%} & \textbf{+0.56\%} & \textbf{+0.34\%} \\
$k_1{=}8$, $k_2{=}20$  & +4.98\% & +4.95\% & +2.18\% \\
$k_1{=}8$, $k_2{=}40$  & +17.18\% & +18.11\% & +6.87\% \\
$k_1{=}8$, $k_2{=}512$ & +128\% & +146\% & +56.04\% \\
\bottomrule
\end{tabular}
\caption{Perplexity relative to the native baseline (absolute PPL in the native rows).
Held-out is a WikiText split disjoint from the split used to choose $k_2$. Perplexity
is minimized at $k_2$ equal to the trained $k$ on both models and all three corpora.}
\label{tab:ppl}
\end{table}

\section{Downstream Validation}

Perplexity on 100K tokens is weak evidence for a claim about model capability. We
therefore validate on MMLU under the paired protocol described earlier
(Table~\ref{tab:mmlu}), and confirm the conclusion on a generation task (GSM8K) and a
Chinese benchmark (C-Eval) below.

\begin{table}[t]
\centering
\small
\setlength{\tabcolsep}{3.5pt}
\begin{tabular}{lrrrr}
\toprule
Configuration & Acc. & $\Delta$\,(pp) & flips $-/+$ & McNemar $p$ \\
\midrule
\multicolumn{5}{l}{\emph{Qwen3.6-35B-A3B}, $n{=}2000$} \\
$k_1{=}8$ native & 81.65\% & --- & --- & --- \\
$k_1{=}6$, $k_2{=}6$   & 79.05\% & --2.60 & 82 / 30 & $9.3\times10^{-7}$ \\
$k_1{=}6$, $k_2{=}8$   & 80.55\% & --1.10 & 53 / 31 & 0.021 \\
\textbf{$k_1{=}6$, $k_2{=}16$} & \textbf{82.40\%} & \textbf{+0.75} & 74 / 89 & \textbf{0.27} \\
$k_1{=}6$, $k_2{=}256$ & 54.20\% & --27.45 & 637 / 88 & $10^{-103}$ \\
\addlinespace
$k_1{=}4$, $k_2{=}4$   & 77.00\% & --4.65 & 166 / 73 & $1.7\times10^{-9}$ \\
$k_1{=}4$, $k_2{=}8$   & 78.55\% & --3.10 & 97 / 35 & $6.4\times10^{-8}$ \\
\textbf{$k_1{=}4$, $k_2{=}16$} & \textbf{81.30\%} & \textbf{--0.35} & 99 / 92 & \textbf{0.66} \\
\midrule
\multicolumn{5}{l}{\emph{Qwen3.5-397B-A17B}, $n{=}2000$} \\
$k_1{=}10$ native & 89.45\% & --- & --- & --- \\
$k_1{=}8$, $k_2{=}8$   & 89.60\% & +0.15 & 23 / 26 & 0.78 \\
$k_1{=}8$, $k_2{=}10$  & 89.00\% & --0.45 & 28 / 19 & 0.24 \\
$k_1{=}8$, $k_2{=}20$  & 88.45\% & --1.00 & 46 / 26 & 0.024 \\
\addlinespace
$k_1{=}5$, $k_2{=}5$   & 87.35\% & --2.10 & 78 / 36 & $1.0\times10^{-4}$ \\
\textbf{$k_1{=}5$, $k_2{=}10$} & \textbf{88.90\%} & \textbf{--0.55} & 42 / 31 & \textbf{0.24} \\
$k_1{=}5$, $k_2{=}20$  & 87.75\% & --1.70 & 64 / 30 & $5.9\times10^{-4}$ \\
$k_1{=}5$, $k_2{=}512$ & 70.90\% & --18.55 & 410 / 39 & $10^{-79}$ \\
\bottomrule
\end{tabular}
\caption{MMLU, 5-shot, greedy, paired McNemar test against each model's native
baseline. ``flips'' counts questions the baseline answered correctly and the variant
did not, versus the reverse. Bold rows halve or nearly halve activated routed experts
and are statistically indistinguishable from the baseline.}
\label{tab:mmlu}
\end{table}

\paragraph{Halving activated experts is free, if the denominator is not halved too.}
Activating 4 of 8 experts on the 35B model costs $4.65$ points under standard
renormalization but $0.35$ at $k_2{=}16$, with a near-symmetric flip distribution
($99/92$); activating 5 of 10 on the 397B model costs $2.10$ points at $k_2{=}k_1$ but
$0.55$ at $k_2{=}10$ ($p{=}0.24$). The reference set recovers $4.30$ and $1.55$ points
at identical compute. A non-significant result is not proof of equivalence---the paired
design only bounds any residual loss below the $0.98$-point resolution at $n{=}2000$.

\paragraph{The optimal $k_2$ is model-dependent, and lies in $[k, 2k]$.}
The 35B model prefers $k_2{=}2k$ at both $k_1$ we tried; the 397B model prefers
$k_2{=}k$; neither prefers $k_2{=}k_1$ once the cut is deep. We therefore do not propose
a universal value but that $k_2$ be scanned over $\{k_1, k, 2k\}$ whenever $k_1$ is
reduced. The consistent finding is the negative one: $k_2{=}k_1$, every
implementation's default, is the worst choice in that grid.

\paragraph{The ablation structure matters.}
Had we evaluated only the baseline and our final candidate---common practice---the 35B
result would read as ``top-4 is simply lossless'' and attribute nothing to the
reference set; only the $k_2{=}k_1$ row shows the reduction is damaging by default.
Conversely, the 397B $k_1{=}8$ block, where the default is already fine ($+0.15$),
shows the opposite: reporting only that row would conclude renormalization never
matters.

\paragraph{Perplexity picks a significantly worse operating point.}
On the 35B model perplexity is minimized at $k_2{=}8$ but MMLU at $k_2{=}16$
(Figure~\ref{fig:s}(a)): the perplexity-optimal setting is significantly worse on MMLU
($-1.10$ points, $p{=}0.021$), and the MMLU-optimal one $+1.81\%$ worse on perplexity.
Since selecting $k_2$ on unlabeled text is exactly the cheap protocol one would reach
for, we regard this as the main methodological finding---compression settings chosen on
perplexity alone should be treated as unvalidated. This sharpens the known pattern that
pruned LLMs retain perplexity while degrading on knowledge
tasks~\citep{jaiswal2024compressing}: here the two metrics are not merely of different
sensitivity but optimized at different points, so perplexity mis-ranks configurations.

\paragraph{Generation: the effect survives autoregressive decoding.}
A single-token multiple-choice score cannot reveal damage that accumulates over a
generated sequence, so we repeat the experiment on GSM8K (Table~\ref{tab:gen}). The
pattern matches MMLU, and is if anything sharper. Reducing to $k_1{=}k/2$ under standard
renormalization costs $6.60$ points (35B) and $4.20$ (397B)---larger than the
corresponding MMLU losses, as expected if a miscalibrated gain compounds token by
token---both highly significant. Anchoring the reference set to the native $k$ erases
the loss: $k_1{=}4$, $k_2{=}16$ costs $0.60$ points (35B, $p{=}0.68$) and $k_1{=}5$,
$k_2{=}10$ costs $0.80$ (397B, $p{=}0.34$), neither detectable.
The truncation rate tracks the accuracy loss closely---it reaches $10.4\%$ at the worst
35B setting and falls to $1.4\%$ once the reference set is corrected---so the default
reduction does not merely change answers but degrades the model into unterminated
outputs.

\begin{table}[t]
\centering
\small
\setlength{\tabcolsep}{4pt}
\begin{tabular}{lrrrr}
\toprule
Configuration & Acc. & $\Delta$\,(pp) & McNemar $p$ & trunc. \\
\midrule
\multicolumn{5}{l}{\emph{Qwen3.6-35B-A3B}, $n{=}500$} \\
$k_1{=}8$ native & 95.00\% & --- & --- & 2.8\% \\
$k_1{=}6$, $k_2{=}6$   & 92.80\% & --2.20 & 0.027 & 6.0\% \\
$k_1{=}6$, $k_2{=}8$   & 95.00\% & +0.00 & 1.00 & 2.2\% \\
$k_1{=}4$, $k_2{=}4$   & 88.40\% & --6.60 & $10^{-6}$ & 10.4\% \\
\textbf{$k_1{=}4$, $k_2{=}16$} & \textbf{94.40\%} & \textbf{--0.60} & \textbf{0.68} & 1.4\% \\
\midrule
\multicolumn{5}{l}{\emph{Qwen3.5-397B-A17B}, $n{=}500$} \\
$k_1{=}10$ native & 96.60\% & --- & --- & 2.6\% \\
$k_1{=}8$, $k_2{=}8$   & 94.60\% & --2.00 & 0.006 & 3.8\% \\
$k_1{=}8$, $k_2{=}10$  & 96.20\% & --0.40 & 0.63 & 2.2\% \\
$k_1{=}5$, $k_2{=}5$   & 92.40\% & --4.20 & $10^{-5}$ & 7.0\% \\
\textbf{$k_1{=}5$, $k_2{=}10$} & \textbf{95.80\%} & \textbf{--0.80} & \textbf{0.34} & 1.6\% \\
\bottomrule
\end{tabular}
\caption{GSM8K, 0-shot, greedy decoding, paired McNemar test against each model's
native baseline. ``trunc.'' is the fraction of generations that reach the 512-token
limit without emitting an answer. Bold rows halve the activated experts with no
detectable loss; the default reduction ($k_2{=}k_1$) is significantly worse and
truncates more often.}
\label{tab:gen}
\end{table}

\paragraph{C-Eval: insensitivity is model-specific, not benchmark-specific.}
On C-Eval (1300 Chinese questions, 5-shot) the 35B model shows no significant
difference for \emph{any} configuration, including the default reduction ($k_1{=}6$,
$k_2{=}6$: $+0.23$, $p{=}0.80$; $k_1{=}4$, $k_2{=}8$: $-0.77$, $p{=}0.31$; baseline
$83.92\%$). Read alone, this benchmark would license the default reduction that MMLU
and GSM8K reject. Yet the same benchmark on the 397B model rejects it: $k_1{=}8$,
$k_2{=}8$ loses $2.92$ points ($p{<}10^{-4}$; baseline $88.31\%$), and anchoring to the
native $k$ recovers most of it ($k_1{=}8$, $k_2{=}10$: $-1.00$; $k_1{=}5$, $k_2{=}10$:
$-0.77$, $p{=}0.20$). The recovery direction is consistent with every other
result---anchoring helps, $k_2{=}k_1$ is worst---but a benchmark's \emph{sensitivity}
to the intervention is not a property of the benchmark alone: the very C-Eval that
certifies the default reduction on one model refutes it on another. Validating on a
single (benchmark, model) pair risks certifying an artifact of that pair---our
perplexity caution, one level up.

\paragraph{Cost accounting.}
On the 35B model, halving $k_1$ from 8 to 4 cuts routed-expert compute in half
($1.006$B to $0.503$B activated routed parameters per token) and the total expert term
by $44\%$ ($1.132$B to $0.629$B), a $17\%$ end-to-end reduction against the $\sim$$3.0$B
parameters per token---though the expert term dominates weight traffic during
memory-bound decoding. Memory is unchanged (all $E$ experts stay resident), so this is a
pure compute-side lever, orthogonal to quantization.

\section{Discussion}

\paragraph{Practical recommendation.}
When compute or decode bandwidth binds, decouple the reference set from the activation
count whenever $k$ is changed and scan $k_2 \in \{k_1, k, 2k\}$ with paired testing:
one integer, three evaluations, and the difference between a significant $2$--$5$ point
MMLU regression and none. It is also a control for any published top-$k$ reduction
result, which absent a decoupled $k_2$ overstates the cost of reducing $k$.

\paragraph{Why we do not recommend expert pruning here.}
Structural compression looks like the harder path on this architecture: the router is
well balanced with essentially no dead experts, retention falls off quickly
(Figure~\ref{fig:routing}(b)), and specialization is domain-bound, so a general-purpose
pruned model has almost no room ($198/256$ experts in the two-domain union). Since the
routed experts are homogeneous and $89.6\%$ of parameters, quantization looks better
matched to the memory problem and composes with our compute-only method. This claim
rests on activation-mass retention and cross-domain transfer; it does not contradict the
near-lossless single-domain pruning of \citet{lasby2025reap}.

\paragraph{Limitations.}
Both models are from the same series; though they differ by $11\times$ in parameters
and in depth, expert count, and trained $k$, cross-architecture generalization remains
a hypothesis. The mechanism predicts the effect scales with router flatness, so
coarse-grained MoEs (e.g.\ 8 experts, top-2) should benefit little; a preliminary check
on a third architecture (Appendix~A) is consistent but ran under an earlier scalar
parameterization and is not directly comparable. Our corpora are English and Python
only, long-context is untested, and the perplexity-optimal $k_2$ is already
domain-dependent. Downstream we cover multiple-choice (MMLU, C-Eval) and short-form
generation (GSM8K) but not long-form generation. Finally, we scan $k_2$ over a
geometric grid; a finer grid, or a per-layer or per-domain $k_2$, may do better.

\section{Conclusion}

Training-free reduction of activated experts in a fine-grained MoE is usually presented
as a compute/accuracy trade-off. We showed it is confounded by an implicit variable:
renormalization applies a gain calibrated to the trained $k$, and lowering $k$ shrinks
the denominator with the activation count, silently miscalibrating the expert branch.
Decoupling the two---activate $k_1$ experts, normalize by the top-$k_2$ mass---turns a
$4.65$ point MMLU regression at half the activated experts into $0.35$ points (35B) and
a $2.10$ point regression into $0.55$ points (397B), for one integer and no measurable
compute. Because perplexity and downstream accuracy select different reference sets,
this knob should be validated on a downstream task with paired testing, not on
perplexity.

\bibliography{refs}

\appendix
\section{Appendix A: A Third Architecture}

Before adopting the $k_1/k_2$ parameterization we ran a preliminary study on
Gemma-4-26B-A4B (30 layers, 128 experts, top-8). It used a global scalar gain
$w_i = \alpha \cdot p_i / m_{k_1}(x)$---a fixed, rather than per-token, version of the
correction studied above---and a different evaluation protocol, so its numbers are not
directly comparable to the main results and we report them as directional evidence only.
They are nonetheless informative on two points: whether a scalar gain helps on a
third architecture, and whether the flatness mechanism predicts the size of the effect.

\paragraph{The scalar gain helps but does not fully recover.}
On MMLU ($n{=}1000$, 5-shot, chat template, instruction-tuned variant, baseline
$80.40\%$), reducing to $k_1{=}6$ at the native $\alpha{=}1$ costs $2.30$ points
(Table~\ref{tab:gemma}). Scanning $\alpha$ recovers up to $1.00$ point: the best value
$\alpha{=}0.80$ reaches $-1.30$ ($p{=}0.079$), the only non-significant setting, but
unlike the two Qwen models it does not return all the way to the baseline. As on Qwen,
the ``natural'' guess $\alpha{=}m_6/m_8{=}0.86$ is not the empirical optimum.

\begin{table}[h]
\centering
\small
\begin{tabular}{lrrr}
\toprule
Configuration & Acc. & $\Delta$ (pp) & McNemar $p$ \\
\midrule
$k_1{=}8$ native & 80.40\% & --- & --- \\
$k_1{=}6$, $\alpha{=}1.00$ & 78.10\% & --2.30 & 0.0018 \\
$k_1{=}6$, $\alpha{=}0.90$ & 78.60\% & --1.80 & 0.015 \\
$k_1{=}6$, $\alpha{=}0.85$ & 78.60\% & --1.80 & 0.020 \\
\textbf{$k_1{=}6$, $\alpha{=}0.80$} & \textbf{79.10\%} & \textbf{--1.30} & \textbf{0.079} \\
$k_1{=}6$, $\alpha{=}0.75$ & 78.80\% & --1.60 & 0.033 \\
\bottomrule
\end{tabular}
\caption{Gemma-4-26B-A4B, MMLU $n{=}1000$, scalar-gain scan (preliminary protocol).
Only $\alpha{=}0.80$ is not significantly worse than the baseline.}
\label{tab:gemma}
\end{table}

\paragraph{The amplitude degree of freedom is untrained on all three models.}
Gemma exposes two learnable scale parameters in its router. The pre-softmax temperature
\texttt{router.scale} was trained hard: it moved from its initialization of $1.0$ to a
near-constant $32.06 \pm 0.02$ across all 30 layers (an effective gain of
$32.06 \cdot 2816^{-1/2} \approx 0.60$ on the RMSNorm-scaled logits). The post-selection
amplitude \texttt{per\_expert\_scale}, by contrast, stayed at its initialization: all
$3840$ values are $1.000 \pm 0.011$, essentially untouched, because it is mathematically
redundant with each expert's \texttt{down\_proj} and the gradient flows to the
larger-magnitude weights instead. The amplitude---the exact quantity our $k_2$
controls---is therefore a degree of freedom that training never optimized on any of the
three models we examined: Qwen exposes no such parameter, Gemma has one but left it at
$1.0$, and GLM-style configurations set it by hand. This strengthens rather than
weakens the main claim: we are not correcting an already-optimized quantity but filling
a gap the training procedure leaves open.

\paragraph{The flatness mechanism predicts the effect size.}
The mechanism in the main text predicts the effect should scale with router flatness.
Gemma's top-8 router mass is $0.302$ versus $0.182$ for Qwen3.6-35B-A3B, i.e.\ a
renormalization gain of $3.31\times$ rather than $5.50\times$. Correspondingly the gain
is far less load-bearing: on the base variant, removing renormalization entirely at
$k_1{=}6$ ($k_2{=}E$) raises WikiText perplexity only from $9.158$ to $9.224$ ($+0.7\%$),
and is in fact \emph{better} than standard renormalization at the same $k_1$ ($9.318$)---
whereas the identical intervention costs $+356\%$ on Qwen. The selection-over-weighting
result also replicates: with the eight activated experts fixed, uniform weights cost
$+34.4\%$ WikiText perplexity, sampling $+218\%$, and random selection $+2436\%$. A
matched $k_1/k_2$ replication on this architecture remains future work.

\paragraph{A protocol caution.}
Gemma is a reasoning model whose base and instruction-tuned variants must be evaluated
differently: the instruction-tuned variant scores WikiText perplexity of $573$ under
plain completion but $9.6$ for the base variant, because bare completion triggers
degenerate repetition. All Gemma perplexity numbers above use the base variant; the
MMLU scan uses the instruction-tuned variant under its chat template. This
protocol dependence is the same one noted in Appendix~B.

\section{Appendix B: Reproducibility Notes}

\paragraph{Infrastructure.}
All experiments use PyTorch with the HuggingFace Transformers library; the method is a
drop-in replacement of the router's \texttt{forward}, and no weights are modified. We
verified that the router computation is numerically identical across Transformers
versions 5.3--5.14 (only variable names changed), reproducing the WikiText baseline
perplexity exactly. Qwen3.6-35B-A3B runs in bf16 on a single 98\,GB accelerator
($\sim$72\,GB resident); Qwen3.5-397B-A17B is sharded across the machine
($\sim$752\,GB in bf16). All routing and downstream configurations are inference-only.

\paragraph{Protocol pitfalls.}
Several protocol details changed our results by more than the effects we set out to
measure, and we record them for others. An initial 0-shot, per-option-loglikelihood
MMLU protocol scored the 35B baseline at $53\%$---implausible for a model of this
size---and standard 5-shot single-letter scoring restored it to $81.65\%$; conclusions
drawn under the first protocol would have been meaningless. The model emits reasoning
traces spontaneously even under plain completion formatting, which exhausts the token
budget on generation tasks unless thinking is explicitly disabled via the chat
template. The released generation config defaults to sampling, which must be disabled
for paired comparison. Evaluation protocol must be matched to model form, a failure
that produced the Gemma perplexity anomaly discussed in Appendix~A and that we
initially misattributed to a framework bug. And, as noted, unpaired testing at
$n{=}2000$ cannot resolve differences below $3\%$, larger than most effects reported
here.

\end{document}